\documentclass[sigconf]{acmart}
\AtBeginDocument{%
  }

\setcopyright{acmlicensed}
\copyrightyear{2025}
\acmYear{2025}
\acmDOI{XXXXXXX.XXXXXXX}

\usepackage{listings}
\usepackage{color,soul}

\begin{document}

\title[Enhancing Learned Knowledge in LoRA Adapters Through Efficient Contrastive Decoding on Ascend NPUs]{Enhancing Learned Knowledge in LoRA Adapters Through Efficient Contrastive Decoding on Ascend NPUs}

\author{Morgan Lindsay Heisler}
\affiliation{%
  \institution{Huawei Technologies Canada}
  \city{Burnaby}
  \state{BC}
  \country{Canada}
}
\email{morgan.lindsay.heisler@huawei.com}

\author{Linzi Xing}
\affiliation{%
  \institution{Huawei Technologies Canada}
  \city{Burnaby}
  \state{BC}
  \country{Canada}
}
\email{linzi.xing@huawei.com}

\author{Ge Shi}
\affiliation{%
  \institution{Huawei Technologies Canada}
  \city{Burnaby}
  \state{BC}
  \country{Canada}
}
\email{ge.shi1@huawei.com}

\author{Hanieh Sadri}
\affiliation{%
  \institution{Huawei Technologies Canada}
  \city{Burnaby}
  \state{BC}
  \country{Canada}
}
\email{hanieh.sadri1@huawei.com}

\author{Gursimran Singh}
\affiliation{%
  \institution{Huawei Technologies Canada}
  \city{Burnaby}
  \state{BC}
  \country{Canada}
}
\email{gursimran.singh1@huawei.com}

\author{Weiwei Zhang}
\affiliation{%
  \institution{Huawei Technologies Canada}
  \city{Toronto}
  \state{ON}
  \country{Canada}
}
\email{weiwei.zhang2@huawei.com}

\author{Tao Ye}
\affiliation{%
  \institution{Huawei Technologies}
  \city{Beijing}
  \country{China}
}
\email{yetao1@huawei.com}

\author{Ying Xiong}
\affiliation{%
  \institution{Huawei Technologies Canada}
  \city{Burnaby}
  \state{BC}
  \country{Canada}
}
\email{ying.xiong2@huawei.com}

\author{Yong Zhang}
\affiliation{%
  \institution{Huawei Technologies Canada}
  \city{Burnaby}
  \state{BC}
  \country{Canada}
}
\email{yong.zhang3@huawei.com}

\author{Zhenan Fan}
\affiliation{%
  \institution{Huawei Technologies Canada}
  \city{Burnaby}
  \state{BC}
  \country{Canada}
}
\email{zhenan.fan1@huawei.com}

\renewcommand{\shortauthors}{Heisler et al.}

\begin{abstract}
Huawei Cloud users leverage LoRA (Low-Rank Adaptation) as an efficient and scalable method to fine-tune and customize large language models (LLMs) for application-specific needs. However, tasks that require complex reasoning or deep contextual understanding are often hindered by biases or interference from the base model when using typical decoding methods like greedy or beam search. These biases can lead to generic or task-agnostic responses from the base model instead of leveraging the LoRA-specific adaptations. In this paper, we introduce Contrastive LoRA Decoding (CoLD), a novel decoding framework designed to maximize the use of task-specific knowledge in LoRA-adapted models, resulting in better downstream performance. CoLD utilizes contrastive decoding by scoring candidate tokens based on the divergence between the probability distributions of a LoRA-adapted expert model and the corresponding base model. This approach prioritizes tokens that better align with the LoRA's learned representations, enhancing performance for specialized tasks. Despite its effectiveness, a naive implementation of CoLD can be computationally expensive. Each decoding step requires evaluating multiple token candidates across both the base and expert models, creating significant inefficiencies. To address this challenge, we developed an optimized kernel for Huawei’s Ascend NPU. CoLD demonstrates consistent performance improvements, achieving up to a 5.54\% increase in task accuracy while reducing end-to-end latency by 28\% compared to greedy decoding. This work offers valuable insights into practical and efficient decoding strategies for fine-tuned LLMs, particularly in resource-constrained environments, and has broad implications for applied data science in both cloud and on-premises settings.
\end{abstract}




\received{07 February 2025}

\maketitle

\section{Introduction}
\label{intro}

In recent years, Low-Rank Adaptation (LoRA) \citep{hu2022lora} has emerged as a powerful and efficient method for fine-tuning large language models (LLMs), gaining significant traction in cloud environments, particularly among Huawei Cloud users. Its ability to enhance pre-trained models without the need for full re-training makes LoRA especially valuable in resource-constrained settings, where efficiency and scalability are critical. 
However, LoRA has been observed to suffer from poor performance on complex tasks, such as chain-of-thought (CoT) reasoning, leaving customers unsatisfied and often resorting to resource-intensive, full-parameter fine-tuning. This significantly limits the practical utility of LoRA, especially in environments where computational accuracy is critical. Previous work has attributed this shortcoming to suboptimal decoding strategies, where interference from the base model undermines the effectiveness of the fine-tuned adapters. Specifically, greedy decoding amplifies inherent biases from the base model, as it tends to favor high-probability, generic tokens over task-specific outputs \cite{shi2024thorough} as seen in Table \ref{tab:qualitative}. Consequently, the LoRA adapters are unable to effectively steer the decoding process, resulting in outputs that are overly generic and misaligned with the specific requirements of the customer’s application. This challenge underscores the need for more sophisticated decoding strategies that can mitigate base model biases and better leverage the LoRA adaptations. 

\begin{figure}[t]
\vskip 0.2in
\begin{center}
\centerline{\includegraphics[width=0.99\linewidth]{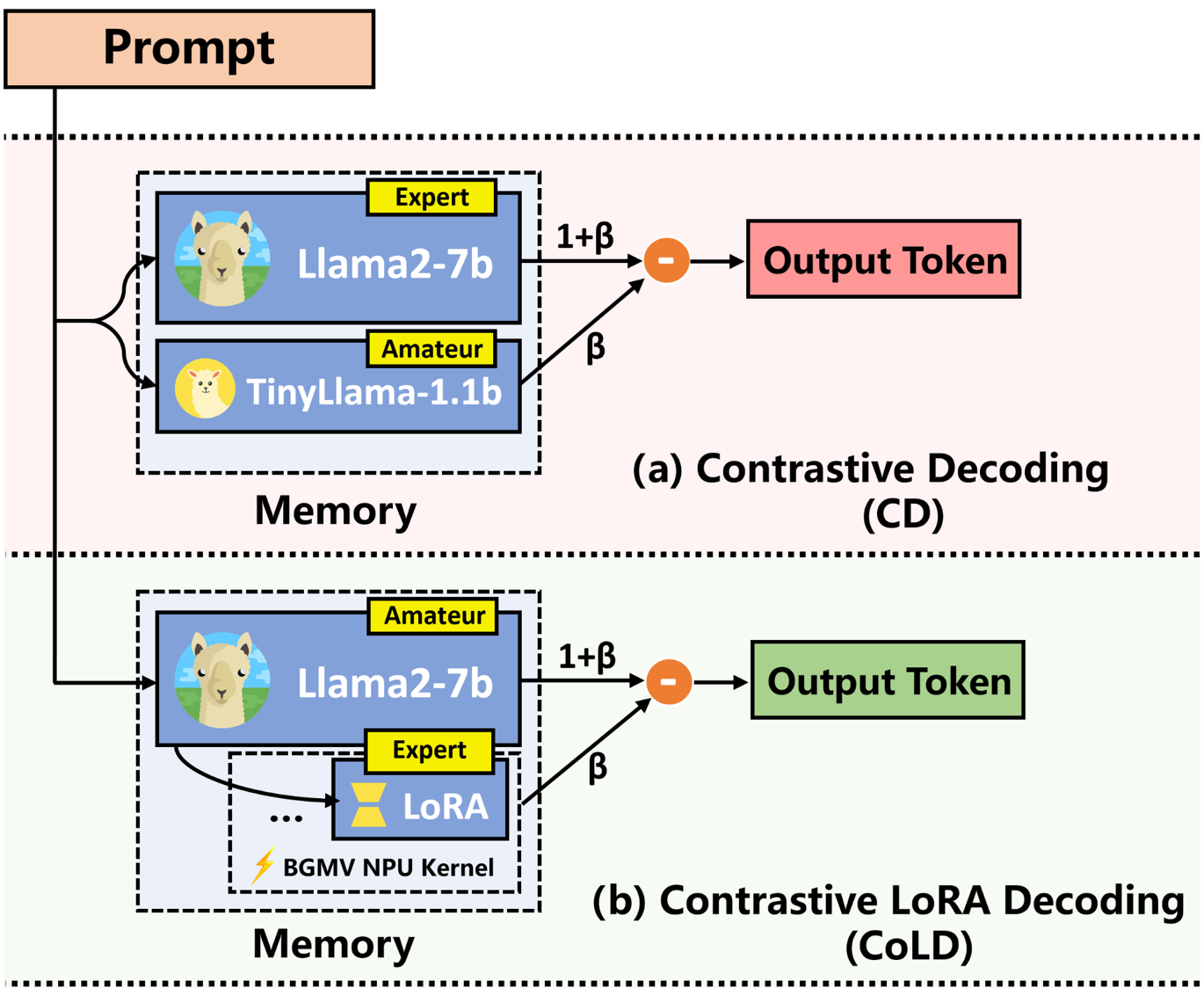}}
\caption{\textbf{Contrastive LoRA Decoding Framework on Huawei Cloud Service}: In practice, contrastive decoding (a) operates through the contrastive interaction between a small amateur model and a large expert model, typically selected from the same model family. Our proposed CoLD (b) formulates the LoRA adapter, combined with the base language model, as the expert model, while the base language model alone serves as the amateur model.}
\label{icml-historical}
\end{center}
\vskip -0.2in
\end{figure}

To address these decoding inefficiencies, we propose a \textbf{Contrastive LoRA Decoding (CoLD)} framework to unlock the full potential of LoRA, delivering high-quality, task-specific outputs without sacrificing computational efficiency as shown in \ref{tab:qualitative}. Our method builds upon contrastive decoding \cite{li2023contrastive, o2023contrastive}—a technique that enhances accuracy by scoring candidate tokens based on the divergence between an expert model's probability distribution and that of an amateur model. Contrastive decoding (as shown in Figure~\ref{icml-historical}(a)) promotes the selection of tokens that align with the expert model's learned representations, making it ideal for tasks requiring enhanced reasoning. In contrast, our proposed CoLD framework (as shown in Figure \ref{icml-historical}(b)) formulates the LoRA-adapted model as the expert model, while the base language model alone serves as the amateur model. This approach can effectively emphasize task-specific signals preserved in the LoRA adapter, thereby enhancing the accuracy of the decoding process. 

\begin{table}[h]
\centering
\small
\begin{tabular}{p{0.18\linewidth}p{0.75\linewidth}}
\toprule
\textbf{Question} & Betty is saving money for a new wallet which costs \$100. Betty has only half of the money she needs. Her parents decided to give her \$15 for that purpose, and her grandparents twice as much as her parents. How much more money does Betty need to buy the wallet?" \\
\midrule
\textbf{Method} & \textbf{Output} \\
\midrule
Greedy & Betty has only half of the money she needs. She has $100 / 2 = 50$ dollars. Her parents gave her $15$ dollars. Her grandparents gave her $15 * 2 = <<15*2=30>>30$ dollars. \textcolor{red}{So Betty has $50 + 30 = <<50+30=80>>80$ dollars.} She still needs $100 - 80 = <<100-80=20>>20$ dollars. \\
\textbf{CoLD} & Betty has only half of the money she needs. She has $100 / 2 = 50$ dollars. Her parents gave her $15$ dollars. Her grandparents gave her $15 * 2 = <<15*2=30>>30$ dollars. So she needs $50 - 15 - 30 = <<50-15-30=5>>5$ more dollars. \textcolor{green}{The answer is $5$.} \checkmark \\
\bottomrule
\end{tabular}
\caption{Qualitative comparison on a CSQA example. CoLD generates more informative and accurate responses.}
\label{tab:qualitative}
\end{table}

However, naively implementing CoLD presents several challenges. First, contrastive decoding inherently introduces computational overhead, as it requires scoring candidate tokens based on the divergence between two models \cite{phan2024distillation}. This results in slower generation speeds compared to greedy decoding, which selects tokens based solely on the highest probability. The additional computations per decoding step can significantly impact latency, making real-time or large-scale deployment impractical without further optimization. Second, CoLD requires both the base and expert models to be loaded into memory simultaneously. This dramatically increases memory consumption compared to greedy decoding, where only a single model is needed. In resource-constrained environments, this added memory requirement can limit scalability and make deployment infeasible on standard hardware configurations. Third, deploying CoLD within Huawei's environment poses unique challenges. Huawei developed Neural Processing Units (NPUs) to accelerate AI workloads and created Ascend-vLLM to optimize model deployment on these NPUs. However, unlike GPUs, which have established libraries like Punica \cite{chen2024punica} for efficient multi-model serving, NPUs do not yet have comparable, off-the-shelf solutions. The architectural differences between GPUs and NPUs—such as memory access patterns and execution pipelines—further complicate efficient model serving. As a result, adapting CoLD for NPUs requires custom kernel design to achieve the same level of performance and efficiency available in GPU-based implementations.

To overcome these challenges, we design an optimized implementation of CoLD tailored for NPUs. To address memory constraints, we leverage multi-LoRA serving, which allows lightweight adapters to be dynamically applied without requiring both models to reside in memory simultaneously. Assuming a 7B model and a 280MB adapter, CoLD consumes around 14.28 GB of memory, compared to 28 GB for two models, resulting in a memory saving of approximately 13.7 GB (48.9\%). Next, given the lack of off-the-shelf solutions for NPUs, we develop a custom kernel that efficiently handles contrastive decoding on Huawei Cloud’s hardware. This kernel is designed to exploit the unique architectural features of NPUs, such as their specialized execution model, enabling efficient execution without the reliance on existing GPU-based solutions. As a result, CoLD achieves up to a 5.54\% increase in task accuracy, while simultaneously reducing end-to-end latency by 28\% compared to traditional greedy decoding.

The primary significance of this work lies not only in improving decoding performance for LoRA-adapted models, but also in overcoming the practical challenges of deploying such models in cloud environments using NPUs. Our approach emphasizes the importance of design decisions tailored to the constraints and capabilities of hardware accelerators like NPUs. These decisions—including the development of a specialized decoding kernel—address key challenges related to memory usage, computational efficiency, and real-time deployment in production environments.

The contributions of this paper are as follows:
\begin{itemize}
    \item We propose CoLD, an efficient framework that improves accuracy on reasoning tasks in LoRAs, boosting the performance of pre-existing LoRA models. Our approach leads to significant accuracy gains, achieving an 5.54\% improvement on GSM8K, highlighting its effectiveness in refining and optimizing LoRA-based models.
    \item We developed a custom kernel to integrate CoLD with Ascend-vLLM, ensuring seamless compatibility. By supporting multi-LoRA inference, this kernel enables CoLD to have a 48.9\% memory saving and 28\% reduction in end-to-end latency.
    \item Our approach has been successfully integrated into a public-facing feature on Huawei Cloud, offering concrete solutions for improving model performance in production environments.  
\end{itemize}

This work aims to bridge the gap between research and real-world applications by presenting an efficient and scalable decoding solution tailored to the needs of industry practitioners.

\section{Related Work}
\label{sec:related}
\noindent\textbf{Decoding Strategies for Large Language Models.} Decoding is a critical step in language generation, determining how predictions from the model’s output distribution are transformed into coherent text. Popular decoding methods include greedy decoding, beam search, and sampling-based approaches \cite{ippolito2019comparison}. While effective for generic LLMs, these methods often fail to adapt to the specific knowledge introduced during fine-tuning. Recent work has explored decoding enhancements, such as nucleus sampling~\cite{Holtzman2020The}, contrastive decoding \cite{li2023contrastive}, and reranking techniques \cite{carraro2024enhancing}. Contrastive decoding, in particular, has shown promise in improving output diversity and consistency by balancing high-confidence and low-confidence tokens. However, these approaches typically target generic LLMs and do not account for the unique characteristics of LoRA-adapted models.
\\~\\
\noindent\textbf{Contrastive Methods in Language Modeling.} Contrastive learning has gained attention in NLP as a means of improving representation learning and output quality. Techniques such as InfoNCE \cite{oord2018representation} and contrastive token-level reranking \cite{yang2024contrastive} have demonstrated significant performance improvements in downstream tasks. Contrastive decoding, recently introduced in \cite{li2023contrastive}, applies contrastive principles to decoding by penalizing low-confidence token sequences while encouraging consistency across predictions. While contrastive decoding has demonstrated improvements in tasks requiring complex reasoning, its reliance on dual model evaluations per decoding step results in substantial computational and memory overhead. To mitigate this, we propose leveraging the unique architecture of LoRAs. 
\\~\\
\textbf{Low-Rank Adaptation in Large Language Models.} Low-Rank Adaptation (LoRA) has emerged as a powerful approach for efficiently fine-tuning large language models (LLMs) by introducing low-rank updates to the model’s weight matrices \cite{hu2022lora}. Unlike traditional fine-tuning methods, which modify all model parameters, LoRA significantly reduces memory and computational costs, enabling its widespread adoption for resource-constrained applications \cite{hayoulora+, chronopoulou2023adaptersoupweightaveragingimprove}. This approach is particularly useful in commercial settings where a single base model must serve multiple use cases. Businesses can adapt quickly by applying different task-specific adapters, which can be done efficiently with multi-LoRA serving. 
\\~\\
\noindent\textbf{Multi-LoRA Serving.} Multi-LoRA serving enables efficient deployment of multiple LoRA adapters alongside a single base model, reducing memory cost by dynamically managing low-rank add-ons instead of full models.
This flexibility also enhances system adaptability by enabling the combination of multiple adapters to tailor models to highly specific requirements \citep{ben-zaken-etal-2022-bitfit, sheng2023s}, further lowering the barriers to AI commercialization.
Additionally, this approach opens the possibility of multi-tenant LoRA serving \cite{chen2024punica}, where adapters from different customers can be served together. This capability improves utilization and reduces costs for cloud service providers, making AI solutions more scalable and cost-effective.

Since requests to individual adapters might on average be sparse, it is critical to batch requests across adapters to improve system throughput and resource utilization.
Naive implementations, however, suffer from inefficiency: the Gather-BMM approach requires gathering adapter weights into contiguous memory before performing batched matrix multiplications (BMM), incurring redundant data copying and unoptimized computational overhead \cite{chen2024punica}.
To address this challenge, Punica introduced the Batched Gather Matrix-Vector Multiplication (BGMV) kernel to optimize memory access and parallelism for multi-LoRA computations \cite{chen2024punica}.
Punica proposed two kernel schedules, BGMV-shrink and BGMV-expand, tailored for LoRA's down-projections (shrinkage) and up-projections (expansion) with distinct parallelization strategies. 
S-LoRA  extended this idea by batching adapters of varying ranks through paged adapter weight management and a rank-aware Multi-size Batched Gather Matrix-Vector Multiplication (MBGMV) kernel \cite{sheng2023s}.
However, existing multi-LoRA serving systems rely on these GPU-specific CUDA kernels, leaving them incompatible with NPUs.
\\~\\
\noindent\textbf{Applications of Efficient LLMs in Industry.} The deployment of LLMs in industry settings often prioritizes efficiency, scalability, and ease of integration. Approaches such as LoRA and adapter-based fine-tuning \cite{pfeiffer2020adapterhub} have facilitated the adoption of LLMs in applications like healthcare \cite{10.1145/3637528.3671576}, marketing \cite{10.1145/3637528.3671583}, commentary generation~\cite{10.1145/3637528.3671537}, and personalized recommendations \cite{parthasarathy2024ultimate}. Despite these advancements, decoding strategies tailored to fine-tuned models have not kept pace, leaving performance gains on reasoning-intensive tasks largely untapped. Our work addresses this gap by introducing a decoding method explicitly designed to enhance the effectiveness of LoRA-adapted models, making them more suitable for real-world applications.

\section{Methodology}
\label{sec:methods}

In this section, we introduce Contrastive LoRA Decoding (CoLD), a novel decoding framework designed to enhance the performance of LoRA-adapted models while maintaining computational efficiency. We first outline the problem formulation, followed by a detailed description of CoLD and its key components, including a novel kernel design for the Ascend-vLLM NPU framework.

\subsection{Problem Formulation}

Let $\mathcal{M}_{\theta}$ be a pre-trained large language model (LLM) with parameters $\theta$. Given a dataset $\mathcal{D} = \{(x_i, y_i)\}_{i=1}^{N}$, where $x_i$ represents an input query and $y_i$ is the corresponding target response, the goal of fine-tuning is to adapt $\mathcal{M}_{\theta}$ to a specific task. LoRA enables this adaptation by learning low-rank updates $\Delta \theta$ while keeping the original model parameters frozen, yielding a task-adapted model $\mathcal{M}_{\theta + \Delta \theta}$.  

During inference, standard decoding strategies (e.g., greedy decoding, beam search, nucleus sampling) operate on the model’s output probability distribution $P(y \mid x; \theta + \Delta \theta)$. However, these methods do not explicitly enhance the knowledge encoded in LoRA adapters, often leading to suboptimal responses in reasoning-intensive tasks. We address this limitation by introducing CoLD, which refines token selection through a contrastive scoring mechanism.

\subsection{Scoring Mechanism}
\label{scoring}

Let \( s_a^i \) and \( s_e^i \) be the unnormalized logits assigned to token \( i \) by the base (amateur) and LoRA-adapted (expert) models, respectively. CoLD leverages these logits to ensure that token selection is guided by the fine-tuned LoRA adapter while avoiding generic, high-probability completions from the base model. There are two key elements in this process: an expert adapter's mask and base model's contrastive penalty.

\subsubsection{Expert Adapter’s \(\alpha\)-Mask}

To refine token selection, CoLD introduces a thresholding mechanism that masks token probabilities based on if their expert logits fall below a dynamic threshold. The threshold is determined by the hyperparameter \( \alpha \), which represents a proportion of the maximum score assigned by the expert model. Tokens that do not meet this threshold are excluded from consideration.

The \(\alpha\)-mask is defined as follows:
\[
V_{\text{valid}} = \{ j \in V \mid s_e^j \geq \log \alpha + \max_{k \in V} s_e^k \}
\]
where \( V_{\text{valid}} \) is the set of valid tokens that pass the \(\alpha\)-mask, and \( V \) is the vocabulary. The value of \( \alpha \) controls the strictness of the mask, with higher values ensuring that only highly probable tokens from the fine-tuned model are retained. This prevents the model from defaulting to generic, high-frequency tokens.

\subsubsection{Base Model’s Contrastive Penalty}

To prevent over-reliance on high-probability tokens from the base model and encourage diversity, the logits from the base model are used to penalize tokens that the base assigns high scores. A penalty is applied using the hyperparameter \( \beta \), which controls the strength of the contrastive penalty. The base model's logits are subtracted from the expert adapter's logits, ensuring that tokens favored by the base model are less likely to be chosen.

The contrastive decoding logits \( s_{\text{CD}}^i \) are computed as:
\[
s_{\text{CD}}^i = 
\begin{cases}
(1 + \beta) s_e^i - \beta s_a^i, & \text{if } i \in V_{\text{valid}} \\
-\infty, & \text{if } i \notin V_{\text{valid}}
\end{cases}
\]
Here, \( (1 + \beta) \) serves as a coefficient that adjusts the scale of the expert logits while keeping the contrastive penalty distinct from the sampling temperature. Tokens outside of the valid set \( V_{\text{valid}} \) are assigned a score of \( -\infty \), ensuring they are excluded from the final output. This filtering step ensures that LoRA-enhanced knowledge is prioritized while preventing generic or low-quality completions.

\subsection{Optimization Objective}

The overall objective of the CoLD framework is to balance accuracy and efficiency by leveraging the strength of the task-specific LoRA adapter. By adjusting the \( \alpha \)-mask and the \( \beta \) penalty, the method provides a flexible approach to open-ended text generation, ensuring that the output remains coherent while avoiding repetitive or overly predictable sequences.

\subsection{NPU Kernel for Multi-LoRA Inference}
\label{sec::kernelmethods}

A key design goal of CoLD is to improve output quality without significantly increasing computational overhead. Unlike contrastive decoding methods that require an additional reference model \cite{li2023contrastive}, CoLD operates directly on the fine-tuned LoRA-adapted model, requiring only minor modifications to the probability distribution during inference. The additional computational cost is limited to small matrix projections and a contrastive calculation per timestep. To achieve this, we developed an optimized kernel on NPU, as discussed in this section, starting with an overview of the NPU architecture and the BGMV kernel schedules for NPU, followed by the details of BGMV shrink and expand schedules.



\subsubsection{NPU Architecture}

Huawei Ascend NPUs are specifically optimized for AI workloads through a hierarchical design. Each NPU integrates multiple AI cores, with each core subdivided into specialized units: (1) \emph{cube} units handle matrix computations, (2) \emph{vector} units perform element-wise arithmetic operations in Single Instruction, Multiple Data (SIMD) fasion, (3) \emph{scalar} units act as compact CPU cores for scalar computations and control flow management, and (4) \emph{memory transfer engine} (MTE) units orchestrate data movement between on-chip buffers and global memory. For a comprehensive discussion of the architectural details of Ascend NPUs, we refer the reader to \cite{liao2021ascend}.

A critical feature of the architecture is the dual role of the scalar unit: it directly executes scalar instructions (e.g., address offset computations, branches and loops) while dispatching non-scalar operations (e.g., cube or vector computations, data movements) to their respective units through corresponding instruction queues. The other units receive instructions and execute them asynchronously unless synchronized via explicit pipeline instructions. As a result, the programming model of Ascend NPUs diverge significantly from that of GPU architectures: unlike CUDA's Single Instruction, Multiple Threads (SIMT) model, Ascend NPUs adopt an asynchronous, pipeline-driven programming paradigm. Consequently, techniques developed for CUDA kernels like Punica \cite{chen2024punica} cannot be adopted directly, necessitating different kernel designs.

Efficient kernel development on Ascend NPUs thus requires careful orchestrations on maximizing pipeline overlap between units (e.g. between vector and cube, or between computing units and MTE) to reduce bubbles. The Ascend C programming framework alleviates developer complexity by automatically optimizing pipeline schedules for many kernels, including strategies like double buffering. This abstraction enables developers to focus on high-level logic while the framework handles low-level pipeline synchronization and resource allocation.


\subsubsection{BGMV Kernel Schedules for NPU}

The LLM inference process can be divided into two stages: \emph{prefill} and \emph{decode}. In the prefill stage, the model processes all batched input tokens. For multi-LoRA computations, we cluster requests by their LoRA adapter indices and perform grouped matrix multiplications using existing NPU kernels. In the decode stage, output tokens are generated autoregressively. Multi-LoRA computations in this stage involve batched matrix-vector multiplications with low operational intensity \cite{chen2024punica}; we therefore designed optimized kernel schedules on the NPU vector units. To avoid confusion in terminology, we refer to a full-width SIMD vector in Ascend NPUs as a \emph{repeat}.



Following vLLM \cite{kwon2023vllm}, the BGMV kernel takes an input tensor $x$, an index tensor $I$, and multiple LoRA weights of the same base model layer from different adapters, where $A_j$ or $B_j$ is from adapter~$j$. For the $i$-th request, two continuous kernel invocations compute the LoRA updates of $x_i$ using the $I_i$-th adapter: $\Delta y_i = x_i A_{I_i} B_{I_i}$. A negative index $I_i=-1$ indicates a base model inference request, which CoLD uses to compute amateur logits.

Both kernel schedules utilize the scalar unit to retrieve LoRA adapter indices. Parallelism between AI cores is achieved across both batch size and output feature dimensions to maximize resource efficiency even when the batch size is small.



Batched matrix-vector multiplications on vector units can be realized via element-wise multiplications followed by horizontal sum reductions. Despite this shared computational paradigm, BGMV-shrink and BGMV-expand have distinct assumptions about their input and output shapes, demanding different implementations.

\subsubsection{BGMV-Shrink} 
BGMV-shrink computes multiple LoRA down-projections of $v_{i}=x_{i}A_{I_{i}}$. In this schedule, the input dimension usually spans multiple repeats but the output dimension (LoRA rank) usually fits within a single repeat. For each output element, we perform an inter-repeat horizontal sum operation over the input dimension, with necessary tiling applied when buffer capacity is exceeded.
However, this horizontal reduction introduces a pipeline dependency: after each reduction, the vector unit stalls until the scalar unit collects the scalar summed result and dispatches data movement instructions to the MTE for the next tile. This dependency serializes vector computations and MTE operations, degrading overall efficiency.
To mitigate this, we leverage software pipelining for the scalar unit to dispatch the MTE operation for the next tile before dispatching the vector computation for the current tile, alleviating the pipeline inefficiency.

\subsubsection{BGMV-Expand} 
BGMV-expand computes multiple LoRA up-projections of $\Delta y_{i}=v_{i}B_{I_i}$.
In this schedule, as the input dimension (LoRA rank) is typically small than a full repeat, we duplicate the input to match the repeat width and use intra-repeat horizontal sum operations to compute output tiles.
This approach avoids interference between vector and scalar units, allowing for automatic double buffering by the Ascend C programming framework.
Additionally, we also fuse the addition back to the base model result into the kernel $y'_{i}=y_{i}+\Delta y_i = x_i W + x_i A_{I_i} B_{I_i}$ to reduce overhead.

\subsection{Implementation and Integration}

CoLD is implemented as a lightweight module within standard transformer-based inference pipelines. It operates as a post-processing step that refines token probabilities before final selection, making it compatible with widely used frameworks such as Hugging Face Transformers for GPUs and Ascend-vLLM for NPUs. The method is hyperparameter-efficient, requiring only $\alpha$ and $\beta$ to be tuned based on task complexity. To assist others in implementing the algorithm effectively and to showcase its simplicity, we provide detailed instructions for implementation within the open-source vLLM framework \cite{kwon2023vllm}, as Ascend-vLLM is close-source.

\subsubsection{API Design.}
\label{app:code}

The main contrastive decoding algorithm, as described in \cite{o2023contrastive}, can be integrated into the vLLM framework by adding it to the \texttt{lora/layers.py} file within the \texttt{\_get\_logits} function definition. Then, we introduce a new data class to define the parameters for contrastive decoding. Add the following code to your \texttt{vllm/sampling\_params.py}:

\begin{lstlisting}[language=Python]
@dataclass
class ContrastiveParams:
    alpha: float = 0.1
    beta: float = 0.5
\end{lstlisting}

\noindent
This class enables users to toggle contrastive decoding on or off and customize its behavior through the \texttt{alpha} and \texttt{beta} parameters.

In the user's main script, the \texttt{ContrastiveParams} class should be imported, and a request can be made as demonstrated below:

\begin{lstlisting}[language=Python]
from vllm.sampling_params import ContrastiveParams

# Example request using CoLD
requests = [
    (
        prompt, 
        SamplingParams(
            temperature=0.0,
            logprobs=1,
            prompt_logprobs=1,
            max_tokens=1024,
            stop_token_ids=[2277],
            (*@\hl{contrastive\_params=ContrastiveParams(}  @*)
                (*@\hl{alpha=alpha,} @*)
                (*@\hl{beta=beta)} @*)
        ),
        LoRARequest("lora", 1, lora_path)
    )
]
\end{lstlisting}



\section{Experimental Methods}

To evaluate the effectiveness of Contrastive LoRA Decoding (CoLD), we conduct experiments across multiple structured and open-ended text generation tasks. We compare our approach against established decoding strategies using both automated evaluation.

\subsection{Datasets}

Consistent with prior work, we use three benchmark datasets, each tailored to specific tasks to provide a comprehensive assessment of CoLD’s ability to refine generation quality across diverse domains. First, for algebraic word problem solving, we utilize the GSM8K (Grade-School Math 8K) dataset, which contains 8,000 math word problems focusing on multi-step reasoning, particularly in algebraic and arithmetic contexts \citep{cobbe2021training}. This dataset is commonly used to assess the model’s ability to break down and solve structured mathematical problems expressed in natural language. For data-to-text generation tasks, we employ the ViGGO (Visual Grounding for Graph-to-Text Generation) dataset, which consists of over 25,000 instances of structured data (graphs) paired with human-written descriptive text \citep{juraska-etal-2019-viggo}. The task evaluates the model’s ability to generate coherent text based on complex, structured data inputs. Finally, for commonsense reasoning, we use the Commonsense Question-Answering (CSQA) dataset, which includes 12,247 multiple-choice questions designed to test the model’s ability to apply general world knowledge and everyday reasoning \citep{talmor-etal-2019-commonsenseqa}. These datasets allow us to comprehensively evaluate our approach across diverse natural language processing tasks.

\subsection{Baseline Methods}

We compare CoLD against standard decoding strategies. Greedy decoding provides a lower-bound reference by selecting the highest-probability token at each step. Beam search ($k=2$) improves upon this by maintaining multiple candidate sequences. Nucleus sampling ($p=0.7$) introduces stochasticity by sampling from a dynamically truncated token distribution. We also include contrastive decoding (CD) \cite{li2023contrastive}, which penalizes high-confidence amateur model outputs to promote diversity. These baselines allow us to evaluate CoLD’s ability to improve output quality while maintaining computational efficiency.

\subsection{Evaluation Metrics}

Performance is assessed using automated evaluation. For the more structured task of Data-to-Text generation, we report BLEU scores \cite{papineni2002bleu} to measure n-gram overlap. For commonsense reasoning and arithmetic tests, we use exact match accuracy to measure the accuracy of the final answer.  
These metrics are chosen to directly evaluate the effectiveness of our decoding methods in tasks requiring high levels of precision and reasoning, ensuring that improvements are both practical and interpretable.

\subsection{Implementation Details}

Experiments are conducted using Llama-2 7B \cite{touvron2023llama} as the base model with LoRA adapters fine-tuned via the PEFT library \cite{hu2022lora}. All LoRAs in this paper are publicly available from previously published work~\cite{chen2024punica, puerto2024dcot}. To ensure fair comparisons, all decoding methods operate at a fixed sampling temperature of 1. CoLD introduces two hyperparameters: $\alpha$, which controls the expert adapter’s filtering threshold, and $\beta$, which modulates the contrastive penalty. We set $\alpha$ and $\beta$ for each task based on validation experiments, to balance precision with adaptability.  

For inference, we optimize execution using Ascend-vLLM. Despite introducing additional filtering and contrastive adjustments, CoLD maintains computational efficiency. Experiments are conducted on a single Huawei Ascend 910B NPU with 65536MB memory. Each experiment uses a batch size of 1, with a maximum output length of 256 tokens.

\begin{table*}[t]
\label{key-results}
\vskip 0.15in
\begin{center}
\begin{small}
\begin{sc}
\begin{tabular}{lcccr}
\toprule
\textbf{Method} & \textbf{Expert Model} & \textbf{Amateur Model} & \textbf{Result} \\
\midrule
\textbf{No LoRA Methods} & & & \\ \midrule
Greedy & - & Llama-2 7B & 14.32 \\
CD \citep{o2023contrastive} & Llama-2 7B & Llama-2 1B & 15.39 \\
DCD \citep{phan2024distillation}& Llama-2 7B & Llama-2 7B + Dropout & 17.28 \\ \midrule
\textbf{LoRA Methods} & & & \\ \midrule
Greedy & LoRA FT Llama-2 7B & - & 27.06 \\
CD  & LoRA FT Llama-2 7B & TinyLlama-1.1B & 28.43 (\textcolor{green}{+1.37}) \\
CoLD (ours) & LoRA FT Llama-2 7B & Llama-2 7B & \textbf{33.06} (\textcolor{green}{+5.54}) \\
\bottomrule
\end{tabular}
\caption{Performance comparison on the GSM8K arithmetic dataset. CoLD achieves the highest accuracy for this task.}
\label{tab:automatic-metrics}
\end{sc}
\end{small}
\end{center}
\vskip -0.1in
\end{table*}

\section{Results}
\label{sec:results}
We begin by showcasing the performance improvements of Contrastive LoRA Decoding (CoLD) over traditional methods on complex reasoning tasks, such as Chain-of-Thought benchmarks, where it achieves a 5.54\% accuracy gain. Next, we evaluate CoLD’s efficiency on NPUs, demonstrating a 28\% reduction in latency compared to greedy decoding, thanks to our optimized kernel within the Ascend-vLLM framework. An experiment highlighting the impact of our kernel optimizations further underscores its effectiveness in resource-constrained Huawei Cloud environments. Additional ablation tests assess the performance across different base models and tasks. Collectively, these results position CoLD as a practical solution for deploying fine-tuned LLMs in industrial applications where both performance and efficiency are critical. Qualitative examples are provided in Appendix \ref{app:qualitative}.



\subsection{Main Results}

Table \ref{tab:automatic-metrics} shows our results within the broader landscape of decoding methods evaluated on the GSM8K dataset, offering insights into advancements over existing techniques.

Contrastive Decoding \citep{o2023contrastive} has demonstrated incremental performance improvements over Greedy Decoding by leveraging amateur models to refine expert outputs. Consistent with prior findings, our implementation of CD achieves an accuracy of 15.39\%, a modest 1.07\% improvement over the baseline Greedy Decoding (14.32\%) using Llama-2 7B. Similarly, Distilled Contrastive Decoding (DCD)~\citep{phan2024distillation}, which incorporates dropout to simulate amateur-like outputs, further enhances accuracy to 17.28\%. These results align with existing literature, which highlights the benefits of refining expert model outputs through contrastive techniques and dropout mechanisms. However, performance remains suboptimal due to the absence of task-specific knowledge.

Next, we evaluate several task-aware LoRA methods. Greedy decoding with a LoRA fine-tuned Llama-2 7B model achieves an accuracy of 27.06\%, highlighting the substantial benefits of targeted fine-tuning. Introducing contrastive decoding (CD) constraints with a smaller amateur model slightly improves performance to 28.43\%, consistent with existing literature showing that contrastive methods yield modest accuracy gains. 

Finally, our proposed approach, that leverages LoRA fine-tuning on both expert and amateur models (Llama-2 7B), achieves a remarkable 
accuracy of 33.06\%. This result surpasses LoRA with greedy decoding by 5.54\%, showcasing the efficacy of jointly optimizing both models to maximize contrastive refinement. 


Interestingly, unlike vanilla contrastive decoding in pretrained models, we find that using a stronger amateur model leads to better performance. Specifically, the Llama-2 7B model achieves a higher accuracy (33.06\%) compared to the smaller TinyLlama-1.1B model (28.43\%). This suggests that refining an expert model by addressing subtle errors is more effective than simply contrasting it with an obviously weaker baseline. By suppressing nuanced mistakes in an already strong model, we can achieve greater performance gains. Instead of focusing on stark differences between expert and amateur models, the goal should be to enhance the expert’s outputs through fine-grained corrections.

\subsection{Kernel Latency} 
A primary goal of CoLD is to improve decoding quality while minimizing computational overhead (see Section \ref{sec::kernelmethods}). To assess this, we conducted an experiment comparing the latency of our kernel against a naive Gather-BMM implementation. The experiment was performed using Llama-2 7B on a single Ascend 910B NPU. For each request, the input consisted of 128 tokens, and 2,048 tokens were generated as output. Each request was paired with a distinct LoRA adapter. We report the average time per output token (TPOT) as our key performance metric.

\begin{figure}[h]
  \centering
  \includegraphics[width=\linewidth]{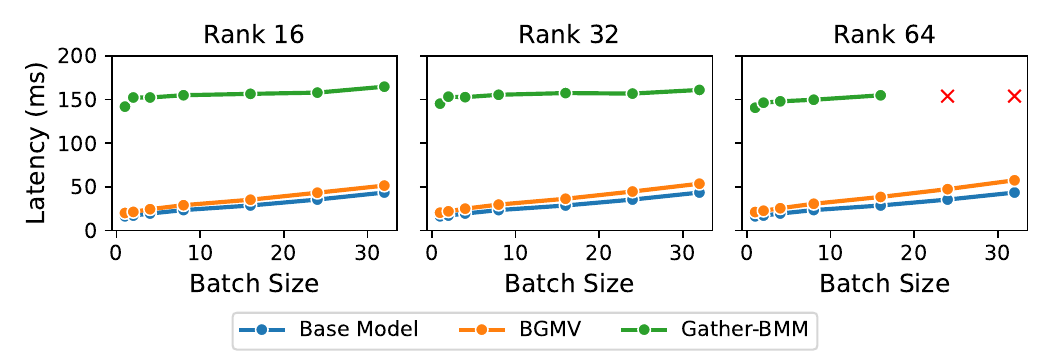}
  \caption{TPOT comparison between Gather-BMM and BGMV on Ascend NPU. Gather-BMM runs out of memory at rank 64 for batch size larger than 16.}
  \label{fig:efficiency}
\end{figure}

The results are shown in Figure \ref{fig:efficiency}. As illustrated, the Gather-BMM method runs out of memory at rank 64 when the batch size exceeds 16, whereas our kernel operates efficiently across all scenarios. This occurs because the Gather-BMM method requires the allocation of intermediate tensors, making it more memory-intensive than our kernel, which avoids the need for additional memory. 
Furthermore, we observed a notable performance slowdown with Gather-BMM relative to our kernel, consistent with trends seen on GPUs \cite{chen2024punica}. In contrast, our kernel shows minimal performance degradation compared to the base model.

\begin{table}[h]
\centering
\small
\begin{tabular}{lcc}
\toprule
\textbf{Method} & \textbf{Output Tokens/s} & \textbf{Average Latency (s)} \\
\midrule
Greedy & \textbf{31.70} & 3.67$\pm$2.97 \\
CoLD (Ours) & 31.20 & \textbf{2.62$\pm$2.12} \\
Nucleus Sampling & 30.50 & 4.05$\pm$2.99 \\
Beam Search ($k=2$) & 14.15 & 3.90$\pm$2.39 \\
\bottomrule
\end{tabular}
\caption{Decoding efficiency comparison with Ascend-vLLM with the first 50 questions on the GSM8K dataset. CoLD operates with a 28.6\% lower average end-to-end latency compared to greedy decoding while significantly improving output quality.}
\label{tab:efficiency}
\end{table}

\begin{table*}[htbp]
\caption{Ablation study showing the zero-shot performance of different decoding methods on the GSM8K dataset with Microsoft's Phi-2 7B models.}
\label{key-results}
\vskip 0.15in
\begin{center}
\begin{small}
\begin{sc}
\begin{tabular}{lcccr}
\toprule
\textbf{Method} & \textbf{Expert Model} & \textbf{Amateur Model} & \textbf{Result} \\
\midrule
Base model Greedy & - & Phi-2 & 27.45 \\
LoRA Greedy & Lora FT Phi-2 & - & 41.18 \\
CoLD (ours) & Lora FT Phi-2 & Phi-2 & \textbf{43.14} (\textcolor{green}{+1.96}) \\
\bottomrule
\end{tabular}
\label{tab:automatic-metrics}
\end{sc}
\end{small}
\end{center}
\vskip -0.1in
\end{table*}

\subsection{Efficiency Analysis}

Building on the kernel efficiency improvements, we now extend the analysis to evaluate the overall efficiency of the CoLD framework. To this end, we provide a comparative efficiency analysis of CoLD versus other decoding methods available on Ascend-vLLM in Table~\ref{tab:efficiency}. With the first 50 questions of the GSM8K dataset, CoLD exhibits lower throughput compared to greedy decoding, but shows a 28.6\% lower average end-to-end latency and better inference speed than both nucleus sampling and beam search. This improved end-to-end latency stems from generating more concise and accurate outputs; unlike greedy decoding, which often produces repetitive and unnecessarily lengthy responses, CoLD reaches the correct answer using fewer tokens on average making it feasible for real-world deployment. The efficiency-cost trade-off is justified by its substantial gains in accuracy and lower end-to-end latency. 


\subsection{Ablation Studies}
We conducted several ablation studies in this study. To see whether our method works on non-Llama models, we provide an example using Microsoft's Phi-2 model in Section \ref{app:phi}. Additionally, we explore different tasks in Sections \ref{CSQA}-\ref{viggo}. Appendix \ref{huggingface} includes an additional experiment comparing performance on Hugging Face and Ascend-vLLM.

\subsubsection{Impact of Base Model Architecture}
\label{app:phi}
We also explore the use of CoLD using non-Llama models. In the following experiments we use Microsoft's Phi-2 model \cite{javaheripi2023phi} with an open-source LoRA adapter \cite{PhiGSM8K}. Phi-2 and Llama-2 7B differ in their design and parameter count. Phi-2 is an instruction-tuned language model with 2.7 billion parameters, optimized for task-specific completions and efficient resource usage. In contrast, Llama-2 7B has 7 billion parameters, designed as a general-purpose model with a broader range of capabilities but requiring significantly more computational resources. As seen in Table \ref{tab:automatic-metrics}, CoLD still  boosts the performance of the pre-existing LoRA model.

\subsubsection{Impact of Task: Commonsense Question Answering}
\label{CSQA}
In our experiments with contrastive decoding \cite{talmor-etal-2019-commonsenseqa}, we observed promising results on the CSQA dataset as shown in Table \ref{tab:text2sql}. By leveraging the contrastive decoding method, we were able to improve the model's reasoning capabilities, as reflected in the higher accuracy scores. The approach facilitated more nuanced understanding and more accurate problem-solving compared to traditional decoding strategies. These findings demonstrate the potential of contrastive decoding in enhancing performance on complex, commonsense reasoning tasks, showcasing its ability to better capture contextual dependencies and improve overall answer correctness in the CSQA dataset. The results highlight the effectiveness of contrastive decoding in handling questions that require deep understanding and reasoning, opening the door for further exploration in similar domains.

\begin{table}[h]
\vskip 0.15in
\begin{center}
\begin{small}
\begin{sc}
\begin{tabular}{lcccr}
\toprule
\textbf{Method} &  \textbf{ViGGO} & \textbf{CSQA} \\
\midrule
Greedy Llama-2 &  20.89 & 32.35 \\
Greedy LoRA  & 31.79 & 49.75\\
CoLD (ours) & \textbf{31.89} (\textcolor{green}{+0.10}) & \textbf{49.87} (\textcolor{green}{+0.12})\\ 
\bottomrule
\end{tabular}
\caption{Performance Comparison on Commonsense Questions and Structured Data Generation}
\label{tab:text2sql}
\end{sc}
\end{small}
\end{center}
\vskip -0.1in
\end{table}

\subsubsection{Impact of Task: Structured Data Generation}
\label{viggo}

In our experiments with data-to-text generation using the Viggo dataset \cite{juraska-etal-2019-viggo}, contrastive decoding showed limited efficacy, with only a 0.10\% BLEU score improvement. This outcome highlights a key limitation of contrastive decoding: its focus on diversity and informativeness is often misaligned with tasks requiring deterministic and highly faithful outputs.\cite{Holtzman2020The} For example, data-to-text generation demands precise mapping from structured data to text, where overly diverse outputs can reduce faithfulness, resulting in negligible performance gains. 


We include these findings to underscore the need to align decoding strategies with task-specific requirements. If the score used for contrastive evaluation doesn't align well with the evaluation metric of a specific downstream task, the method may fail to produce optimal outputs. For example, in tasks where fluency or coherence is more critical than diversity, the prioritization of contrast can mislead the decoding process. For tasks like data-to-text generation, methods prioritizing faithfulness and logical consistency, such adaptive contrastive decoding \cite{zhu2024improving} or adaptive contrastive search~\cite{arias2024adaptivecontrastivesearchuncertaintyguided} (which would also work within the CoLD framework) could improve the applicability of contrastive decoding for such deterministic tasks. Future work could look into adapting these decoding schemes into the framework.



\section{Conclusion}


In this paper, we introduced Contrastive LoRA Decoding (CoLD), a novel decoding framework designed to enhance the performance of LoRA-adapted large language models (LLMs) by efficiently leveraging fine-tuned knowledge during inference. While traditional decoding strategies such as greedy decoding and beam search fail to fully utilize the contextual adaptations introduced by LoRA, our approach addresses this gap by incorporating contrastive principles to improve reasoning and output quality. 

However, contrastive decoding methods are often computationally intensive and memory-hungry, making them impractical for deployment in real-world systems, especially on NPUs. To overcome this challenge, we developed optimized kernel implementations within the Ascend-vLLM framework on Huawei Cloud, enabling efficient execution of CoLD on NPUs. Our solution achieves significant improvements in both performance and efficiency, with up to a 5.54\% increase in task accuracy and a 28\% reduction in end-to-end latency compared to traditional decoding methods.

This work underscores the importance of tailoring decoding strategies to the hardware and frameworks used in deployment. By bridging the gap between advanced decoding algorithms and efficient inference on NPUs, we offer practical insights for applying fine-tuned LLMs in resource-constrained, industrial settings. 

\bibliographystyle{ACM-Reference-Format}
\bibliography{sample-base}

\appendix

\section{Hugging Face vs. Ascend-vLLM Performance Results}
\label{huggingface}

To demonstrate how much this optimized kernel influences CoLD, we provide a computational efficiency analysis where we observe distinct trade-offs between Hugging Face (unoptimized) and Ascend-vLLM (optimized) inference. On NPUs, CoLD runs inefficiently with Hugging Face as it does not have an optimized multi-LoRA framework. However, in real-world deployment on NPUs via vLLM, inference speeds improved from 9.70 tokens/sec to 31.20 tokens/sec due to optimized memory utilization and parallelized execution. The performance gains on Ascend-vLLM comes with some trade-offs in customization, as Ascend-vLLM requires specific integration steps to maintain compatibility with different transformer architectures.

Table \ref{tab:gpu_npu_comparison} summarizes the key performance metrics of CoLD across Hugging Face and Ascend-vLLM environments, highlighting inference speed, latency, and deployment constraints on the same NPU.

\begin{table}[h]
\centering
\small
\begin{tabular}{|c|c|c|}
\hline
\textbf{Metric} & \textbf{Hugging Face} & \textbf{Ascend-vLLM} \\
\hline
Inference Speed (tokens/sec) & 9.70 & 31.20 \\
Average Latency (s) & 13.06 & 2.62 \\
Deployment Overhead & Low & Moderate \\
Scalability & Moderate & High \\
\hline
\end{tabular}
\caption{Comparison of CoLD performance on Hugging Face and Ascend-vLLM environments.}
\label{tab:gpu_npu_comparison}
\end{table}

Overall, these results indicate that while Hugging Face remains a versatile platform for research and prototyping, Ascend-vLLM offer superior efficiency in production settings.

\section{Qualitative Analysis}
\label{app:qualitative}

An example output from the base Llama-7B model, LoRA Fine-Tuned model, and CoLD from the CSQA dataset is presented in Table \ref{tab:qualitative-examples-csqa}. The base model provided generic factual statements without contextual reasoning, failing to adequately address the specific question. In contrast, the LoRA Fine-Tuned model demonstrated an improvement by organizing and evaluating the answer options more systematically, though it still misjudged the most appropriate response, suggesting "formal garden" rather than the more contextually relevant "garden center." The performance of CoLD outshone both previous models, offering a thorough, context-aware analysis and precise reasoning for each option. This resulted in the correct answer, "garden center," being identified as the most suitable choice. Overall, while the LoRA Fine-Tuned model provides notable progress over the base model, CoLD shows the most accurate, context-sensitive response, demonstrating superior performance in both logical reasoning and task comprehension.


\begin{table*}[htbp]
\centering
\caption{Example outputs from different decoding methods on the CSQA dataset.}
\begin{tabular}{p{3cm}p{10cm}}
\hline
\textbf{Method} & \textbf{Generated Text Example} \\ \hline
Question & Question: If I did not have a rosebush, where would I get one?

Choices: A) pot, B) museum, C) garden center, D) formal garden, E) backyard \\ \hline
Base Model & 1. If I did not have a rosebush, I would have to get one from a garden center or a nursery.

2. A rosebush is a type of plant that is grown in a garden or a backyard.

3. A rosebush is not a type of plant that is found in a museum or a formal garden.

4. A rosebush is not a type of plant that is found in a pot.

5. A rosebush is not a type of plant that is found in a backyard.

6. A rosebush is a type of plant that is found in a garden center or a nursery. \\ \hline
Lora FT & 1. If I did not have a rosebush, where would I get one?

2. A) Pot - This option does not make sense as a rosebush is not grown in a pot.

3. B) Museum - This option is also not relevant as a rosebush is not displayed in a museum.

4. \textcolor{red}{C) Formal garden - This option is possible as a rosebush can be grown in a formal garden.}

5. D) Backyard - This option is the most likely answer as a rosebush can be grown in a backyard.

6. E) Formal garden - This option is not relevant as a rosebush is not grown in a formal garden.

Therefore, the answer is C) formal garden.

[Final answer] C) \textcolor{red}{formal garden} \\ \hline
\textbf{CoLD (ours)} & 1. If I did not have a rosebush: This means I currently have a rosebush or I previously had one.

2. Where would I get one: This means where can I obtain a rosebush.

3. Options: A) pot - a container to grow a rosebush in 

B) museum - unlikely as museums do not sell plants 

\textcolor{green}{C) garden center - a place that sells plants and gardening supplies }

D) formal garden - a type of garden design with structured layout and ornamental plants 

E) backyard - the area behind or surrounding a house or building 

4. Answer: C) garden center - the most likely option as garden centers sell plants and tools for gardening. 

Therefore, if I did not have a rosebush, I would get one from a garden center.

[Final answer] \textcolor{green}{C) garden center} \checkmark \\ \hline
\end{tabular}
\label{tab:qualitative-examples-csqa}
\end{table*}






\end{document}